\documentclass{report}
\usepackage{graphicx} 
\usepackage{hyperref}
\usepackage{amsmath}
\usepackage{dsfont}
\usepackage[numbers]{natbib}

\title{Teaching Robot Policies to Humans Using Erroneous Examples}
\author{\begin{tabular}{c}
    Rithika Narayan\\[14.5pt]
    \textbf{Advisors: } Professor Henny Admoni, Dr. Suresh Kumaar Jayaraman
\end{tabular}
}
\date{May 2025}

\begin{document}

\maketitle

\begin{abstract}
Human-robot collaboration describes the process of humans and autonomous agents working together to accomplish common goals. This process is facilitated best when robot policies, or behaviors in different situations, are made transparent to humans. Demonstration-based explanations have been a focus of human-robot collaboration research, and the field has frequently drawn upon literature from education to improve how humans are taught robot policies. However, no single teaching method has been proven effective across domains, difficulties, learners, and other variables; the question of how humans can most effectively be taught robot policies remains open.

In traditional classrooms, learners are shown erroneous examples, in which they reflect on and correct incorrect responses to understand common pitfalls when learning a concept. We propose using erroneous examples to teach robot policies, extending an existing policy teaching framework. We conduct a user study in which participants view incorrect demonstrations of robot behavior and correct the actions to align with the actual policy. Our findings suggest that viewing these incorrect demonstrations and verbalizing one's reasoning in predicting a robot's actions improves retention of the policy over time, in agreement with the effect of erroneous examples in classrooms. We also categorize participants into distinct learning styles and establish that participants using inverse reinforcement learning-like reasoning perform best on policy prediction tasks. With this work, we aim to advance the methods by which robots educate humans on their policies. 
\end{abstract}

\chapter{Introduction}
As the capabilities of autonomous agents improve and technology advances, humans and autonomous robots are increasingly interacting and collaborating on a diverse set of tasks. It is now commonplace to have a robotic vacuum clean your floors or to work alongside pick-and-place mobile robots in warehouses. In the near future, your surgeon may utilize an autonomous robot to complete your sutures. In all of these environments, humans must be able to understand how to best utilize robots for task completion, and trust robots to perform their tasks efficiently while also protecting the safety of all involved.  

Explainability and transparency are key pillars for increasing human trust in autonomy; if people are able to understand how autonomous agents have reached their decisions, they may feel more comfortable in using these agents \cite{endsley}. For example, many machine learning models used in radiology include heat maps to show which parts of the X-ray are being considered when making a diagnosis. This prevents the deployment of models incorrectly biased by training data and allows radiologists to verify diagnoses \cite{xray}. While image-based models often use saliency maps as explainability tools, other sequential decision-making models may provide natural language explanations or demonstrations of their behavior. The modality of an explanation changes based on the context, but the goal is always to promote human understanding of autonomous agents.

In addition to trust, explainability and transparency have implications on task efficiency and safety. In collaborative assembly tasks, robots often are programmed to stop when a human enters their workspace for safety reasons; humans anticipating how the robots will act using knowledge of their decision-making process moderates the number of stops the robots have to take for safety \cite{booth}. 

 A robot's policy is the set of rules governing the actions it takes in different situations. For example, a robot vacuum's policy may include reversing when it encounters stairs or vacuuming around heavy objects. An autonomous vehicle's policy should include stopping when pedestrians are crossing the road and staying within the boundaries of its current lane. Reinforcement learning-based agents, which will be discussed in more detail in following chapters, have policies guided by reward functions. These functions assign rewards and costs to different states and actions. The tradeoffs between rewards and costs inform the robot's decisions. For example, a robot vacuum may be rewarded heavily for picking up dirt and penalized heavily for falling down stairs; the tradeoff between the two may encourage a conservative path when the robot is navigating to dirt from an area near the stairs.  When humans use familiar technologies like robot vacuums, they often have a preconceived notion of how these robots will behave (their policies). In more unfamiliar contexts or high-stakes scenarios in which mutual understanding between humans and autonomous agents is critical, humans must be taught the robot policies explicitly as their intuition may not be enough to accurately predict the robot's actions.

 Prior work has explored explanations through demonstrations of the robot's behavior in different environments \cite{lee1, lee3}. This style of example-based explanation is particularly useful to convey more complex internal reward functions, as compared to mathematically defining the function or providing verbal explanations \cite{shtein}. The demonstrations have been augmented with testing of humans' understanding of the robot and feedback to correct the humans' beliefs. Much of the work in this space has been motivated by findings from educational literature, applying the techniques through which humans learn from each other in the classroom to humans learning from robots. 
 
\begin{figure}[ht]
    \centering
    \includegraphics[
    width=\linewidth]{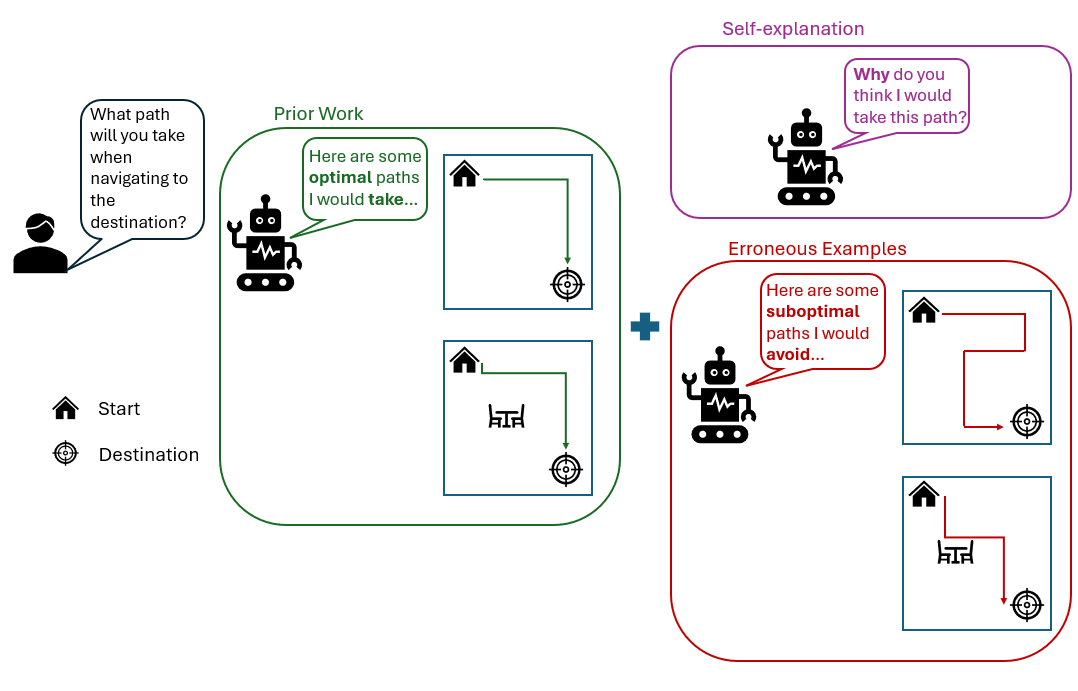}
    \caption{Prior work communicates robot policies via a series of demonstrations of a robot’s optimal trajectories in various environments \cite{lee3}. We propose integrating self-explanations, or verbalization of reasoning, and erroneous examples, or suboptimal trajectories, into the paradigm to improve a human learner’s ability to predict robot trajectories in new environments.}
    \label{fig:fig1}
\end{figure}

 However, prior works still find it difficult to convey complex tasks through demonstrations; there appears to be a ceiling on how well people understand nuanced policies, even after applying multiple interventions \cite{lee3}. One reason could be that these works demonstrate only optimal robot behaviors, i.e. show only positive examples which by itself maybe difficult to comprehend. In mathematics pedagogy, educators have found success in teaching both simple and complex concepts by integrating practices of self-explanation, or verbalizing one's reasoning, and inspection of erroneous examples, or incorrect solutions to prompts \cite{grosse, mclaren, selfex1, selfex2}. Based on these practices employed in traditional classrooms, we propose augmenting  demonstrations and tests of robot policies with  self-explanations, or reflections on beliefs about a robot's policy, and erroneous examples, or demonstrations of suboptimal behavior for the robot in different scenarios, to strengthen the human learner's understanding of the robot policy (Figure \ref{fig:fig1}).

We intend to enhance human understanding of robot decision-making process using erroneous/sub-optimal robot behavioral demonstrations and self-explanations, in order to take advantage of proven benefits these tools have shown in classrooms. 
Our contributions are as follows: 
\begin{itemize}
    \item Integration of self-explanation and erroneous examples into the robot policy teaching loop
    \item A user study exploring the effectiveness of self-explanations and erroneous examples on human ability to predict robot policy
    \item Characterization of human learners of robot policies into learning styles and validation of assumptions made in human belief modeling. 
\end{itemize}

\chapter{Related Works}
We summarize prior works in the fields of explainable reinforcement learning and mathematical pedagogy. The techniques described below are heavily reused and adapted for our work and provide the necessary technical background.

\section{Explainable Reinforcement Learning}
\begin{figure}[ht]
    \centering
    \includegraphics[height = 8 cm]{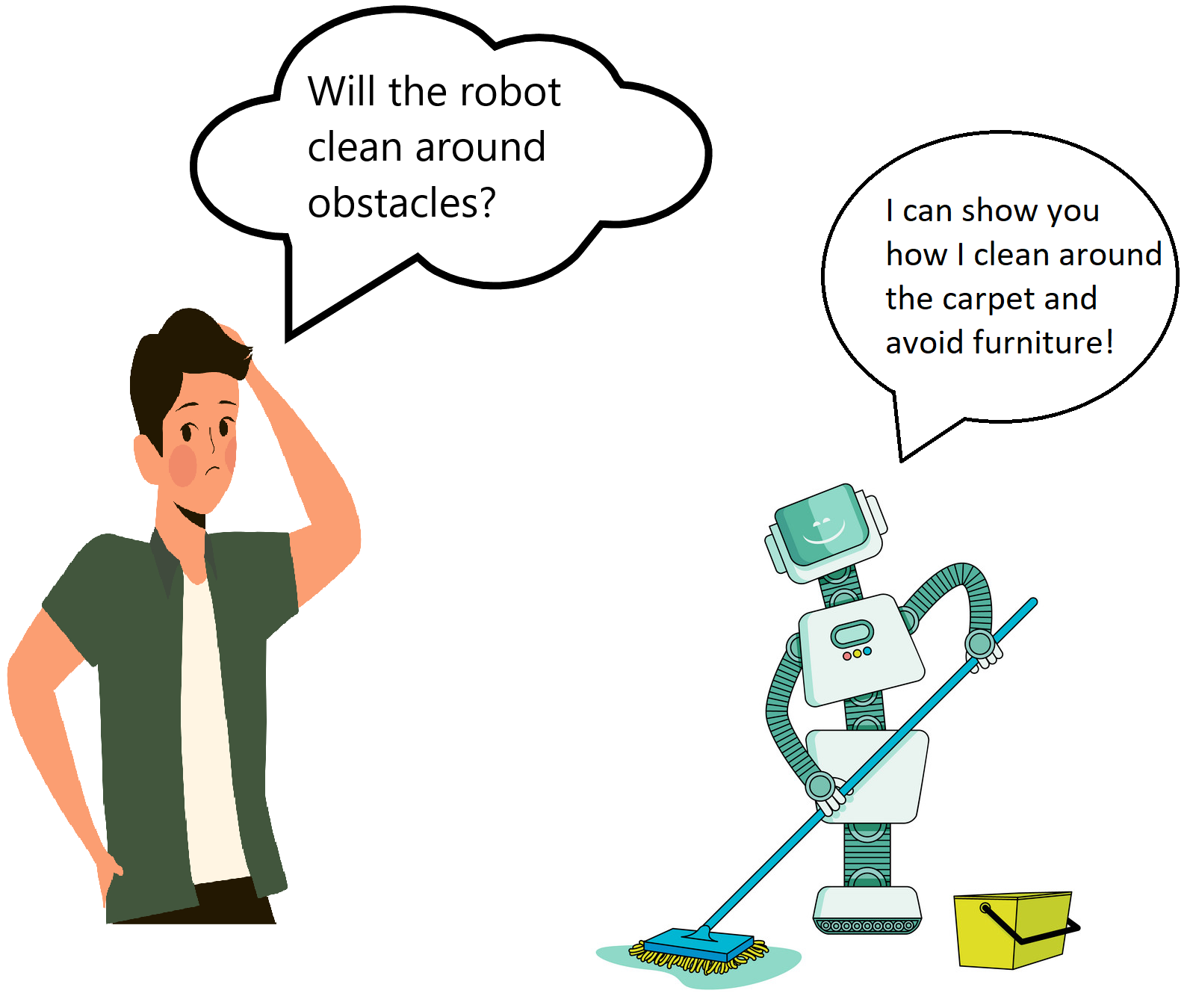}
    \caption{Interaction facilitating a demonstration-style explanation of the agent's behavior}
    \label{fig:clean}
\end{figure}

Explainable reinforcement learning (RL) seeks to make the policies of RL-based agents understandable to humans. RL agents' policies are determined by interacting with the environment to receive rewards and pay penalties, and by learning from these benefits and costs to improve behavior over time. Prior work has sought to make RL agents explainable by directly conveying their numerical rewards and costs to humans; this has been found to be overwhelming and non-intuitive for complex policies and environments \cite{shtein}. Other researchers have explored using inherently interpretable models such as decision trees to approximate the RL model being explained or have used saliency maps to characterize the importance of features in a state \cite{silva, greydanus}. Recent work has explored demonstrations, or example-based explanations, as a different way to convey policies to humans  (Figure \ref{fig:clean}) \cite{lee2}.

Researchers have borrowed from principles learned in the classroom to ``scaffold'' demonstrations that incrementally increase in difficulty to ease humans into learning robot policies \cite{lee1}. Use of scaffolding is beneficial when a policy elicits complex behavior or is influenced by several features in the environment; learners can make progress towards understanding the full policy without becoming overwhelmed by its overall complexity. 
Recent work has focused on incorporating remedial education, which corrects misconceptions students gained during the initial lesson, into a demonstration-based policy teaching loop \cite{lee2}. Students are shown demonstrations of the policy, periodically take tests on the lessons they have learned thus far, and may receive additional demonstrations and tests if they answer incorrectly. We adopt the demonstration-based explanation as the explanation style of interest in our work due to its success in conveying policies of varying complexity. The teaching loop explored in this previous work forms the basis of the framework we propose and is discussed further in the Methods section.

\section{Markov Decision Processes}
In our user study, participants are taught the policies for two robots which are modeled as Markov Decision Processes (MDPs). Each MDP is comprised of a state space $\mathcal{S}$, action space $\mathcal{A}$, transition function $T$, reward function $R$, discount factor $\gamma$, and initial state distribution $\mathcal{S}_1$. When two MDPs share the same $R, \mathcal{A},$ and $\gamma$ they are said to be in the same domain. These domains differ in $T, \mathcal{S},$ and $\mathcal{S}_1$. In a domain, an optimal trajectory for a robot to take in an environment is described as a sequence of initial state, action, and next state tuples $(s_i, a, s_{i + 1})$ that are obtained by following the robot's optimal policy, called $\pi^*$. The robot's reward function $R$ is represented as a weighted linear combination of reward features, $\phi$. The weights assigned to each feature are organized in the vector $\boldsymbol{w}*$, which means we can write $R = \boldsymbol{w}*^\top \phi(s_i, a, s_{i + 1})$.

\section{Machine Teaching for Policies}
As in previous studies teaching robot policies to human learners, we draw from the machine teaching framework set forth by Lage et al. \cite{lage}. In their work, they summarize a robot's policy by  selecting a set of demonstrations of the policy that are informative to the learner and help them recover the robot's optimal policy. We maintain assumptions from prior work that the human's learning process can be modeled through inverse reinforcement learning (IRL) \cite{jara}. The human learner uses the demonstrations to approximate $\boldsymbol{w}*$ and applies planning to predict the robot's behavior within the MDP. The underlying approach to determine the effectiveness of a demonstration in conveying $\boldsymbol{w}*$ utilizes behavioral equivalence classes (BECs). 
 
\section{Behavior Equivalence Classes}
For a given demonstration $\xi^*$, the BEC for that demonstration is the set of possible robot policies for which the demonstration would be optimal. We have formulated our reward functions as weighted linear combinations of features in the environment. The BEC of demonstration $\xi^*$ under optimal policy $\pi^*$ is defined as the half-space formed by the IRL equation:
\[\text{BEC}(\xi^* | \pi^*, \pi_{\boldsymbol{w}}) = \boldsymbol{w}*^\top (\mu_{\pi^*}^s - \mu_{\pi_{\boldsymbol{w}}}^s) \ge 0, s = \xi^*(0)\]
where $\mu_{\pi}^s = \mathds{E}[\sum_{t = 0}^\infty \gamma^t \phi(s_t) | \pi, s_0 = s]$ \cite{lee2022, ng}. This means that $\mu_{\pi}^s$ is the vector of reward feature counts accumulated from starting in state $s$  and following policy $\pi$. $\xi^*(0)$ is the first state of demonstration $\xi^*$. The above equation is used to convert any demonstration into a constraint on the reward vector $\boldsymbol{w}$. The constraint produced by the equation can also be referred to as the knowledge component taught in the demonstration, which captures some part of the reward function such as the relationship between two features \cite{koedinger}. For example, in a robot vacuum's reward function, one knowledge component might be the tradeoff between taking a longer path to the destination and distance from furniture along that path. 

\section{Learning from Self-Explanations in Educational Literature}
As discussed in the Explainable Reinforcement Learning section, explainable RL researchers have been turning to findings from literature on teaching learners in traditional classroom settings to improve methods by which robots teach humans their policies. Drawing on this trend, we discuss findings from teaching sciences to students across education levels. A growing body of work suggests that the practice of self-explanation, or verbalizing one's reasoning in arriving at a solution to a problem, improves reasoning and concept retention \cite{selfex1, selfex2, selfex3}. Chamberland et al. and Chi et al. integrate self-explanation into the studying process for medical students and middle school science students, while Larsen et al. integrate self-explanation into the testing process for medical students \cite{selfex1, selfex2, selfex3}. All three studies asked learners to provide a self-explanation at specific points in the learning task, but did not provide any structure for doing so and collected open-ended, text-based responses. According to McLaren et al., asking for template-based or multiple-choice explanations may be confusing or too cognitively taxing for students to benefit from \cite{mclaren}. Overall, findings across subjects indicate that self-explanation is a useful tool in building overall critical thinking skills and in building comprehension of specific subjects. 

\section{Learning from Erroneous Examples in Educational Literature}
In the last decade, educational researchers have been investigating the value of explaining and correcting incorrect answers. This means that when teaching trigonometry, the teacher might show an example on the board where they flip the use of sine and cosine then have students explain the error and the correct answer. When this technique has been studied in classrooms across age groups, students exposed to incorrect examples tend to show better long-term ability to solve problems \cite{grosse}. McLaren et al. conducted a study on middle school students learning about decimals, in which the control group was asked to work through a word problem after the teacher taught a lesson and the experimental group was shown a word problem with an incorrect answer and asked to explain and correct it. They found that the experimental group performed better on a delayed post-test, suggesting that erroneous examples are beneficial for long-term information gain \cite{mclaren}. In order to generate erroneous examples to show to students, McLaren et al. identified common misconceptions students held (such as decimals with more digits must be greater than decimals with fewer digits, $.9 < .34$) from literature on teaching decimals. They then created incorrect solutions to word problems (the erroneous examples) by working through the problem with reasoning derived from the misconception.

From literature in teaching mathematics, we find that exposing students to erroneous examples and prompting them to critically evaluate the misconceptions improves concept retention in the long term \cite{grosse, mclaren, errex}. Thus, we intend to build a policy teaching framework that incorporates erroneous examples of robot behavior and test whether the benefits of erroneous examples in mathematical education transfer to teaching humans about robot policies. 

While many researchers have tackled the challenge of teaching robot policies to humans, none have ``solved'' this issue. Years of work have pointed the field towards adapting techniques from educational research. Our ability to teach people policies has improved, but it is not a closed question. Thus, we intend to investigate new avenues of teaching robot policies to human stakeholders,
based on promising data regarding the value of self-explanation and erroneous examples in the classroom learning process.

\chapter{Methods}
We begin our methodology with a discussion of the modeling techniques used by a robot teacher to approximate the human learner's understanding of its policy. We then describe the framework by which we adapt educational literature to explain policies to humans and end with a description of the user study conducted to understand the effects of the techniques we propose. 
\section{Particle Filter}

Previous studies have developed a particle filter-based method for modeling human beliefs over the possible reward weights for a given policy, in order to show learners the most effective demonstration based on their current beliefs \cite{lee2, lee3}. Each particle represents a belief in a particular combination of reward weights. The particles undergo a Bayesian update based on the constraints added to the belief space (see the section on Behavior Equivalence Classes) following demonstrations and unit tests (discussed in the following section). The constraints added following demonstrations are expected constraints that the learner has acquired on the belief space; the constraints added following unit tests represent actual knowledge gain that the learner is showing in test performance. A given demonstration may produce multiple expected constraints by comparing the optimal trajectory shown in the demonstration with counterfactual trajectories (which are trajectories that the human may have incorrectly believed that the robot would take). 

\begin{figure}[ht]
    \centering
    \includegraphics[width = \linewidth]{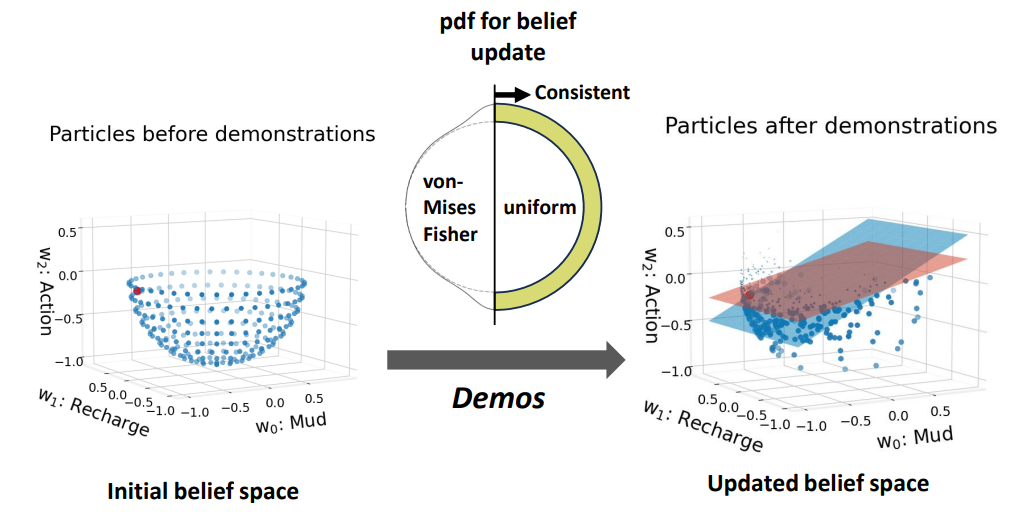}
    \caption{Update of particle filter based on introduction of red and blue half-space constraints in demonstration. Particles consistent with the constraints receive more weight (yellow ring) while particles inconsistent with the demonstrations are weighted with the von-Mises demonstration. \cite{suresh}}
    \label{fig:partial}
\end{figure} 

All constraints are half-space constraints, meaning that they split the belief space equally into one side of beliefs that are consistent with the demonstration or test (meaning that these beliefs would have resulted in the trajectory being shown) and inconsistent beliefs (that would not have resulted in the behavior shown). Prior work has utilized a custom probability distribution function to update particle weights based on the probability distribution enforced by the most recently added constraint. This custom function is comprised of a uniform distribution for particle weight updates on the consistent half-space of the constraint and a von Mises-Fischer distribution over the inconsistent half-space. The von Mises-Fischer distribution asserts that as a particle's distance from the constraint increases, the belief represented in that particle has an exponentially less likelihood of resulting in the behavior shown to generate the constraint. 
This particle filter can be used to sample human beliefs over the reward weights in order to generate a set of demonstrations and tests that are best suited to constrain the human learner's potential ideas of the robot policy. The method for doing so is described in \cite{lee3}. We use this particle filter to generate a fixed curriculum for the user study which all participants experience in order to maintain consistency between this work and prior works. This fixed curriculum was generated by sampling from the particle filter based on scaffolding, or iteratively increasing the difficulty of knowledge components presented to learners, and on counterfactual reasoning, which means that we present demonstrations that are likely to target misconceptions held by learners as a group \cite{lee3}. While we use a fixed curriculum, the particle filter can also be adaptively sampled to generate demonstrations tailored to a particular learner's current beliefs in order to correct their individual misconceptions.  

\section{Robot Policy Teaching Loop}
As alluded to above, the knowledge components in a robot's policy are conveyed to human learners through demonstrations, unit tests, and feedback. We adopt a partial feedback loop first put forth by Lee et al., in which learners see demonstrations and take unit tests periodically and are provided feedback on their unit tests \cite{lee3}. Demonstrations are examples of the robot's optimal trajectory in a new environment, visualized action-by-action for the human learner. Unit tests in prior work are opportunities for the learner to demonstrate their understanding of the policy; they are shown a new environment configuration and asked to indicate the optimal trajectory for the robot in the environment. If the learner provides the optimal trajectory in the unit test, feedback simply tells them that they are correct and may move on. If the learner answers incorrectly, the feedback shows them the optimal trajectory alongside the response they provided. The partial feedback loop is visualized below (Figure \ref{fig:partial}). The loop is repeated until the robot has taught all the knowledge components of the policy.

\begin{figure}[ht]
    \centering
    \includegraphics[height = 3.5 cm]{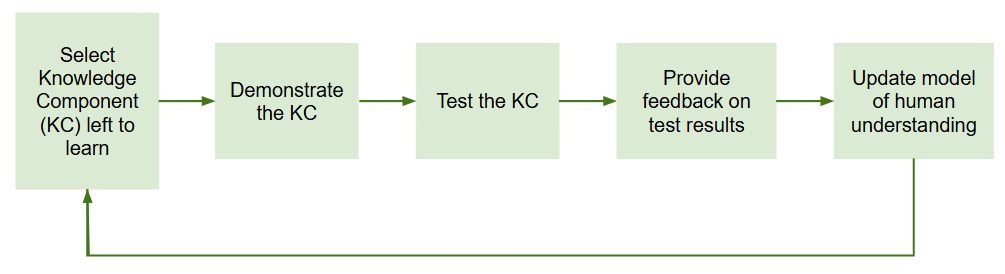}
    \caption{Partial feedback policy teaching loop used in previous work \cite{lee2}}
    \label{fig:partial}
\end{figure} 

\section{User Study}
In order to explore the effects of self-explanation and exposure to erroneous examples on humans' ability to predict robot policy, we conducted a user study in which participants were taught two robot policies and assessed on their ability to predict the robots' actions in unseen environments. 

\subsection{Domains}
Participants were taught the policy of a robot in two domains \cite{lee1}. The policies are formulated as MDPs in a grid world with a finite set of actions with associated action costs and reward features. \\

\textbf{Taxi domain. } In this domain, a taxi robot must pick up a package and deliver it to a destination in a four by three grid world. The actions available are \textit{up, down, left, right, pick up, drop} and each action has a cost of one. The robot is rewarded for dropping off the package at the destination and for passing through recharge stations, and is penalized for traveling through mud.  \\

\textbf{Skateboard domain. } In this domain, a robot must reach a destination in a six by five grid world. There is also sometimes a skateboard present in the environment; the robot is penalized less per action if it is riding the skateboad. The available actions are \textit{up, down, left, right, pick up} and each action has a cost of one (unless the robot is on the skateboard). \\

In the user study, domains were visually masked (ex. using yellow squares instead of mud and green hexagons instead of recharge stations in the taxi domain) in order to avoid biasing participants based on priors. 

\subsection{Study Design}
As discussed in the Robot Policy Teaching Loop section, previous user studies teaching robot policies to humans have developed a policy teaching loop comprised of demonstrations, unit tests, and feedback \cite{lee2}. We adapt this teaching loop to conduct a user study in which there are three distinct treatment groups that receive different formats of unit tests. The unit test format a participant receives will be compared as a between-subjects condition. 

\begin{figure}[ht]
    \centering
    \includegraphics[width = \linewidth]{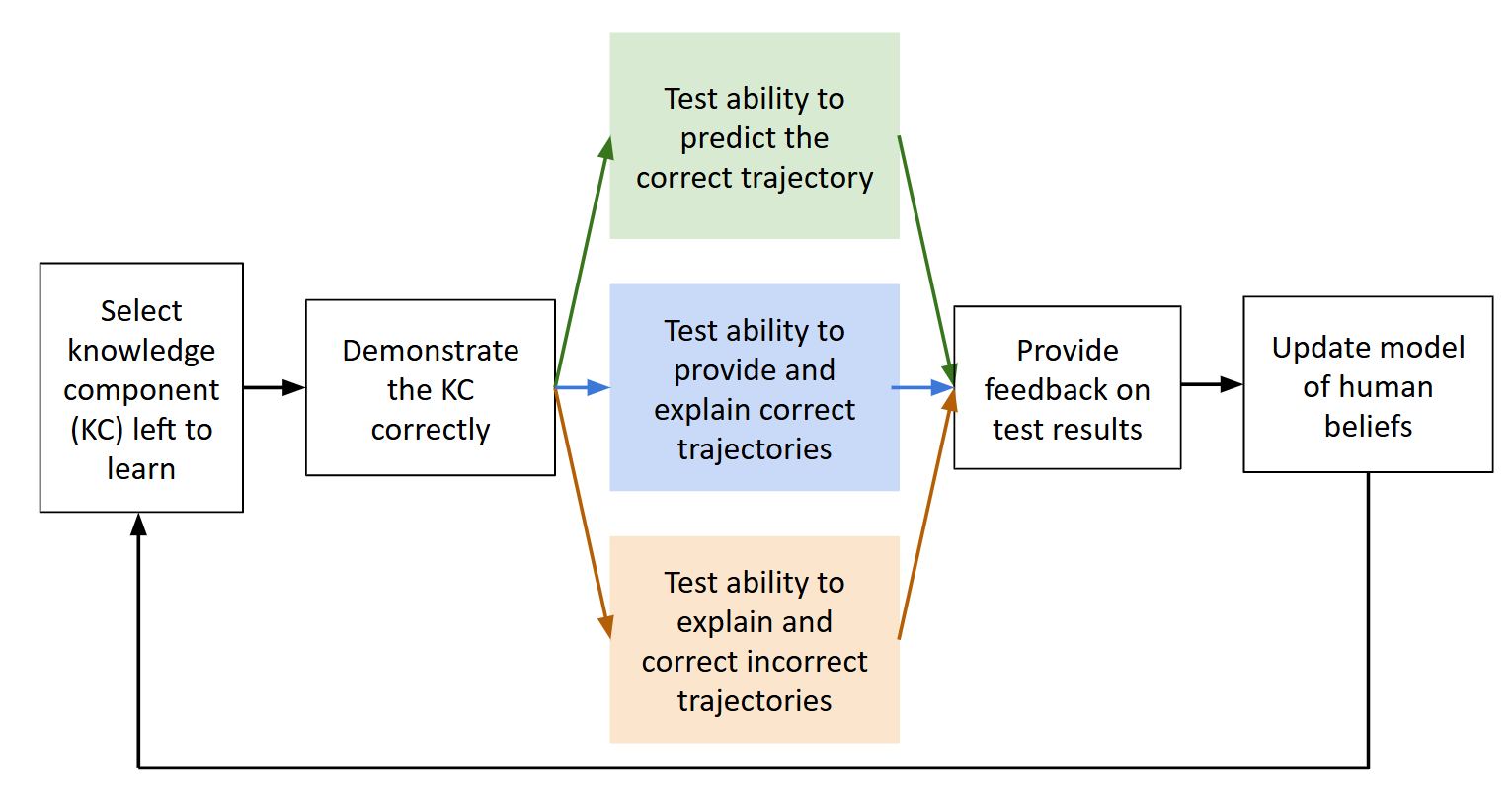}
    \caption{Policy teaching loop with control condition (green), explanation condition (blue), and erroneous example condition (orange)}
    \label{fig:mine}
\end{figure} 

Below, we will show examples of the unit test formats using the taxi domain. However, participants in the study are taught the policies for both the taxi and skateboard domains, in a randomized order. Since all participants will be taught policies for both domains, the effect of domain on participants' ability to predict robot policy will be compared as a within-subjects condition.

All participants experienced an identical teaching curriculum (set of demonstrations and tests) regardless of treatment group. In other words, for a given unit test, a participant in any treatment group will be presented with the same new environment configuration. The difference between treatment groups lies in the prompt about this new environment configuration and the response format. 

\begin{figure}[ht]
    \centering
    \includegraphics[width = \linewidth]{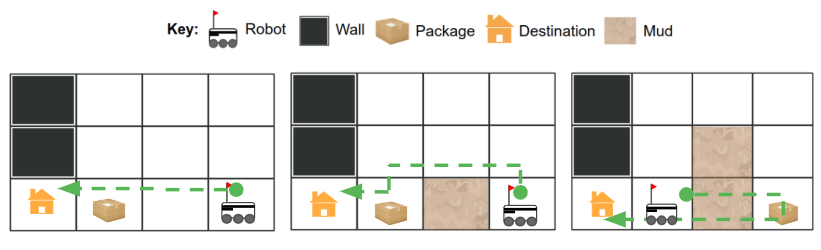}
    \caption{Possible series of demonstrations shown to participants to teach tradeoff of action cost and traveling through mud in taxi domain. }
    \label{fig:demos}
\end{figure}

Participants are first shown demonstrations of the robot's trajectory in sample environments and are then given unseen environments in the tests, as described below. An example of a series of demonstrations teaching the tradeoff between action cost and cost of traveling through mud is shown below (Figure \ref{fig:demos}). 

\begin{figure}[ht]
    \centering
    \includegraphics[width = \linewidth]{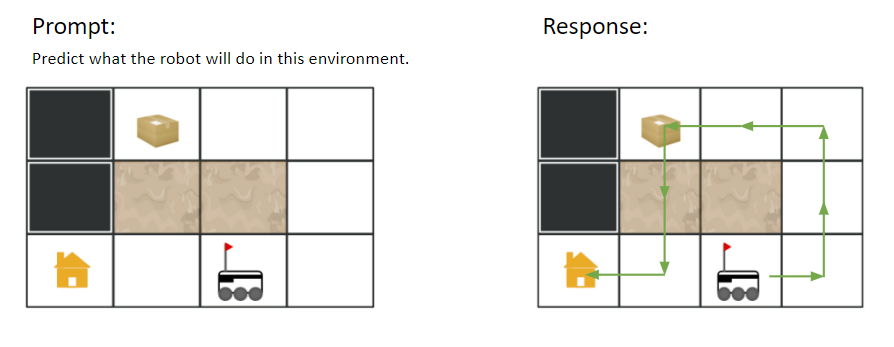}
    \caption{Unit test format under control (green) condition with prompt environment and response trajectory}
    \label{fig:control}
\end{figure}

The first group, the control group, will experience the green testing condition in the training loop in Figure \ref{fig:mine}. After receiving demonstrations on a knowledge component such as the cost of going through mud, control group participants see the robot in a new environment and are asked to predict its behavior in this environment, given what they have learned so far (Figure \ref{fig:control}). This style of unit test is identical to unit tests formats in prior work and is used as a benchmark against which we will examine the effectiveness of the following two treatments. 

\begin{figure}[ht]
    \centering
    \includegraphics[width = \linewidth]{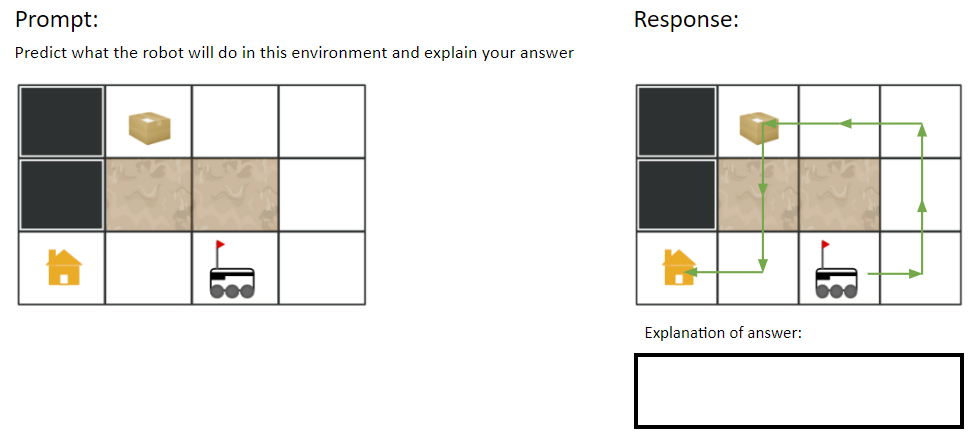}
    \caption{Unit test under second experimental condition with prompt environment, response trajectory, and open-ended explanation}
    \label{fig:group2}
\end{figure}

The second group experiences the self-explanation condition, shown in blue condition in Figure \ref{fig:mine}. In this condition, participants are asked to predict the optimal trajectory in a new environment and must also explain their reasoning in a text box, as seen in Figure \ref{fig:group2}. As discussed in Section 2.5, verbalizing one's reasoning during problem-solving tasks is known to improve performance across many educational levels and disciplines. The role of self-explanation in learning robot policy through demonstrations has not been established in literature, so this condition is included as a bridge to the erroneous condition and provides a novel contribution in itself. 

The third and final group experiences the erroneous example condition, seen in orange in Figure \ref{fig:mine}. Their unit tests take the form of an incorrect trajectory for the robot to take in a new environment; the participant must explain why the response is incorrect through a text box and then provide and explain the correct trajectory. We have included the ability to explain the error and one's own reasoning separately to mirror the prompt and response set up used in educational literature on erroneous examples \cite{mclaren}. As observed in piloting, some learners submitted identical responses for both explanations, some separated their reasoning between explanations, and others used a combination of approaches. 

\begin{figure}[ht]
    \centering
    \includegraphics[width = \linewidth]{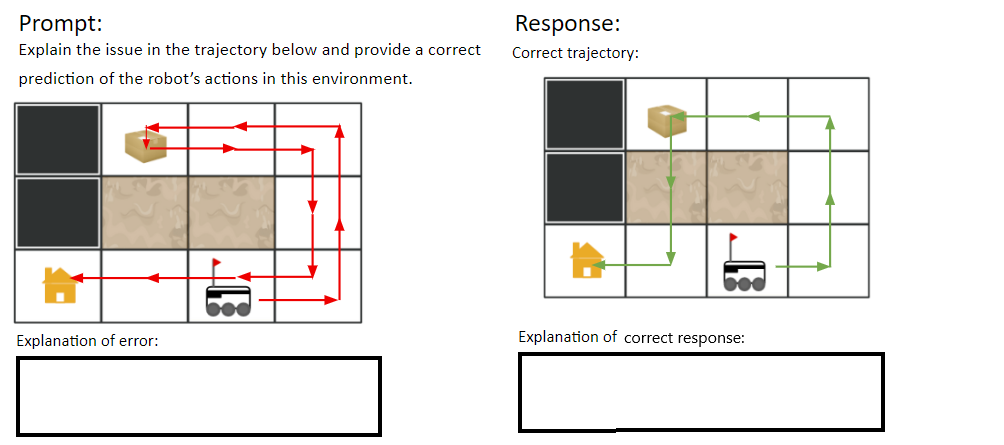}
    \caption{Unit test under third experimental condition, including erroneous example of trajectory, explanation of error, corrected trajectory, and explanation of reasoning}
    \label{fig:group3}
\end{figure} 

If a participant in any treatment group answers a unit test incorrectly, the participant receives feedback which contrasts the learner's incorrect test response with the optimal response in the environment. 

\begin{figure}[ht]
    \centering
    \includegraphics[width = \linewidth]{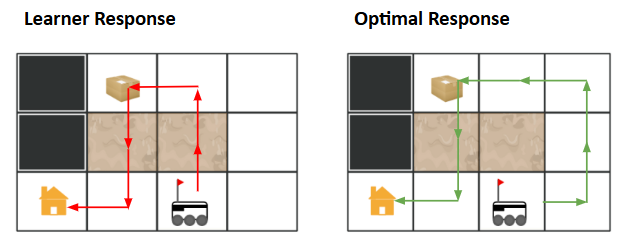}
    \caption{Example of feedback provided to participants contrasting a learner's incorrect test response with the optimal response.}
    \label{fig:feedback}
\end{figure} 

After all knowledge components of the reward function are taught through the policy teaching loop, all participants will take an immediate post-test and a delayed post-test 24 to 36 hours later in which they will be tested across knowledge components. The questions in these tests will take the same form as the prompts posed to the control group: predicting the robot's actions in an unseen environment. We diverge from other studies in which robot teachers convey their policies to human learners by including this delayed post-test. Its purpose is to measure retention of the policy and its inclusion mirrors the assessments used in educational literature to gauge the effect of erroneous examples \cite{mclaren, errex}. Most often, the benefits of erroneous examples do not appear in immediate performance assessments and information gain; instead the benefit becomes apparent in concept retention.

\subsection{Erroneous Example Selection}
In this section, we will discuss how the erroneous examples shown to participants in the third treatment group were selected. 

The curriculum used in our study is the same as the curriculum used by Lee et al., as discussed in the Particle Filter section \cite{lee3}. This was done in order to take advantage of a known, tested curriculum that applied findings from literature to select the most appropriate demonstrations and tests across the learner pool. This was also done in order to facilitate the selection of erroneous examples for this study. 

In educational work studying the effects of erroneous examples, study designers selected erroneous examples that intentionally targeted misconceptions that many students had about the topic being taught. Thus, we identified common misconceptions held by participants in previous studies teaching the policies of the taxi and skateboard domains using the same curriculum. 

From the complete set of participant data obtained by Lee et al., we isolated incorrect responses (trajectories) for each unit test individually  \cite{lee3}.
For a given unit test, if there was a singular trajectory that formed the majority of the incorrect responses to the test, we used this trajectory as the erroneous example. Most of the participants who erred on the given test erred in that specific way, meaning that it was the dominant misconception. 

\begin{figure}[ht]
    \centering
    \includegraphics[width = \linewidth]{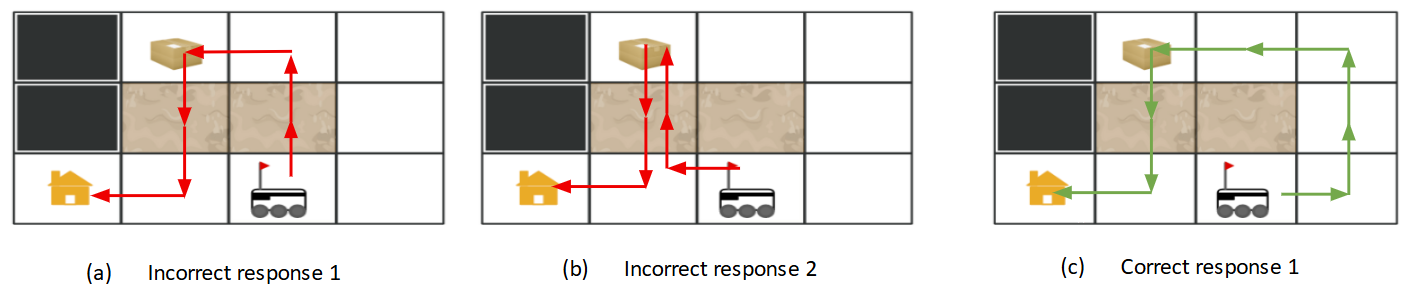}
    \caption{Example of two incorrect test responses ((a) and (b)) that reveal the same underlying misconception compared to the optimal response (c).}
    \label{fig:misconception}
\end{figure}

However, there were some unit tests in which there was not a majority incorrect trajectory. In this case, we identified the dominant misconception overall by examining the reward of each incorrect trajectory and identifying a majority of trajectories which could have resulted from the same incorrect belief of the learner over the belief space and resulted in the same reward over the trajectory. For example, in Figure \ref{fig:misconception}, both trajectories in the taxi domain have the same overall reward (as both took an equal number of actions and traveled through equal numbers of mud squares) and result from the same misconception that the cost of mud is less than twice the cost of taking any action. This misconception causes learners to believe it is worth it to cut through the mud to reach the package, when the optimal trajectory would avoid the mud on the path to the package. Together, the two incorrect responses shown in Figure \ref{fig:misconception}
represented the majority of incorrect responses to the test and the more frequently appearing of the two could be selected as the best candidate for the erroneous example. 

\subsection{Measures}
We recorded the following measures to extract results from the study: \\

\textbf{M1. Optimal Response: } Participants  are assigned a binary score depending on the optimality of their test trajectory. \\

\textbf{M2. Regret: } Participants are assigned a real-valued score based on the difference between the total reward/cost incurred in the optimal trajectory and the total reward/cost incurred by their test trajectory. \\

\textbf{M3. Improved Understanding Rating: } 5-point Likert scale with prompt “Did this [demonstration or test] improve your understanding of game strategy?”\\

\textbf{M4. Cognitive Load: } 5-point Likert scale with prompt “Learning the game strategy was taxing.”\\

The Likert scale for M3 were provided after each demonstration and unit test or feedback interaction. The Likert scale for M4 was provided following the completion of the demonstration, unit test, and feedback section, prior to the post-test. 

\subsection{Hypotheses}
We have six hypotheses corresponding to the above three measures: \\

\textbf{H1. } The erroneous example condition will result in the highest rate of optimal responses on the post-tests and the control condition will result in the lowest. \\

\textbf{H2. } The erroneous example condition will result in the lowest regret across post-test responses and the control condition will result in the highest. \\

\textbf{H3. } The erroneous example condition will result in the greatest improvement/least decline in optimal response between the immediate and delayed post-test, and the control condition will result in the least/greatest, respectively. \\

\textbf{H4. } The erroneous example condition will result in the greatest decrease/least increase in regret between the immediate and delayed post-test, and the control condition will result in the least/greatest, respectively. \\

\textbf{H5. } The erroneous example condition will result in the highest average rating of improved understanding on unit test interactions and the control condition will result in the lowest. \\

\textbf{H6. } The erroneous example condition will result in the highest cognitive load on participants and the control condition will result in the lowest.

\chapter{Results}
We collected data from 39 participants on Prolific \footnote{\hyperlink{https://www.prolific.com/}{https://www.prolific.com/}} for part one of the study, involving the demonstrations, unit tests, feedback, and immediate post-test. 30 of these participants returned to take the delayed post-test 24 to 36 hours following the immediate post-test. Treatment conditions were randomly assigned, as was the order in which the two domains were presented. There were 14 participants in the control group in part one of the study, 13 in the self-explanation group, and 12 in the erroneous explanation group. 44\% of participants were male, 54\% were female, and 2\% chose not to disclose. Ages ranged from 21 to 77 (M = 41.23, SD = 14.90). \\
In part two of the study, 12 members of the control group returned, 10 members of the erroneous example group returned, and 8 members of the self-explanation group returned. 37\% of participants were male, 60\% were female, and 3\% chose not to disclose. Ages ranged from 26 to 77 (M = 43.47, SD = 14.23).

\section{Performance in Part One and Part Two}
\textbf{H1} is not supported. A mixed ANOVA did not find a significant effect of test condition on number of optimal responses on the immediate post-test ($F(1, 39) = 0.65, p = 0.528$). However, there was a significant interaction between domain and number of optimal responses ($F(1, 39) = 6.47 , p = 0.015$). Participants performed much better on the taxi domain, as expected. The tradeoffs between different feature costs/rewards was far tighter in the skateboard domain than in the taxi domain, making it more difficult to learn. Similarly, on the delayed post-test there was no significant effect of test condition on number of optimal responses ($F(1, 30) = 0.136, p = 0.873$). However, there was no significant effect of domain on performance in this second post-test ($F(1, 30) = 1.702, p = 0.202$). We believe that this may be caused by a ceiling on performance in the taxi domain and floor on performance in the skateboard domain in the first post-test; if participants did very well on taxi domain tests initially, there is little room to improve. Similarly, there is little room to worsen performance on the skateboard domain if participants performed poorly on it initially. 

We also chose to inspect regret separately from optimal response, following precedent in literature \cite{lee3}. Prior works have found that optimal response may be too coarse of a metric to accurately assess learning of robot policies; while participants may not provide the correct trajectory in a given test, the differences between the answers they provided and the optimal response are significant and can be indicative of their understanding. \textbf{H2} is also not supported. A mixed ANOVA found no significant effect of test condition on the sum of the regret over the immediate post-test ($F(1, 39) = 0.367, p = 0.695$). Nor did it find a significant effect in the delayed post-test ($F(1, 30) = 0.076, p = 0.927$). As with the optimal response metric, there was a significant effect of domain on the sum of regret in the immediate post-test ($F(1, 39) = 21.423, p < 0.001$). There was also a significant effect of domain on the sum of regret in the delayed post-test ($F(1, 30) = 8.335, p = 0.007$). Participants demonstrated lower regret in the taxi domain, as compared to the delivery domain in  both post-tests.

\section{Change Between Immediate and Delayed Post-tests}
\begin{figure}[!htbp]
    \centering
    \includegraphics[height = 7 cm]{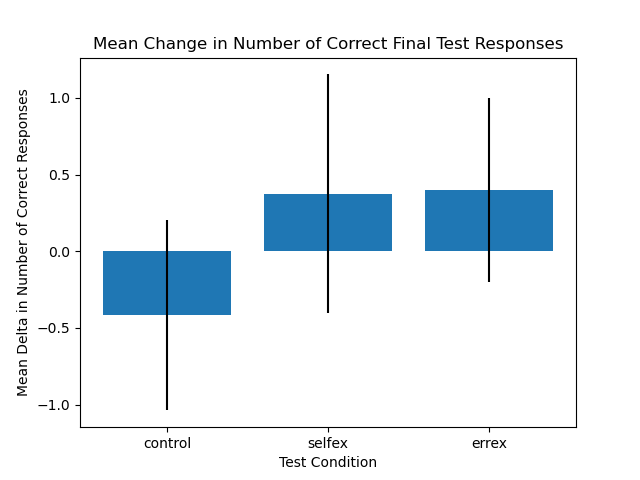}
    \caption{The control group is the only group to answer fewer questions correctly on the delayed post-test than on the immediate post-test. The self-explanation and erroneous example groups show improvement in performance.}
    \label{fig:overallbar}
\end{figure}

\begin{figure}[!htbp]
    \centering
    \includegraphics[height = 7 cm]{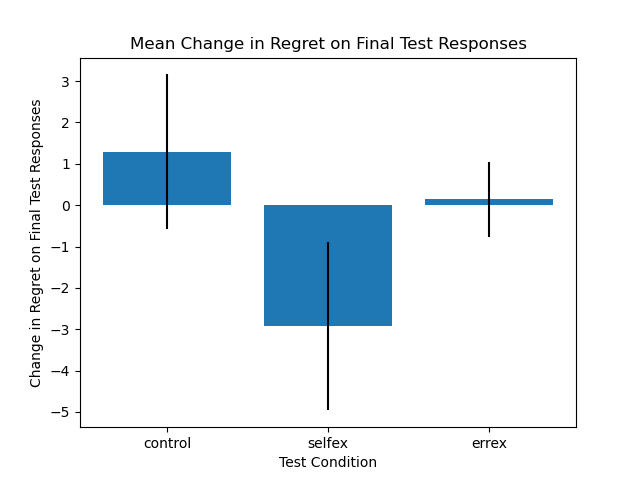}
    \caption{The self-explanation group's average regret decreases greatly from immediate to delayed post-test. This means that their trajectories deviated less from the optimal ones as time passed. }
    \label{fig:regbar}
\end{figure}

\textbf{H3} and \textbf{H4} were not supported, however there are interesting trends in the change in optimal response and regret between the three treatment groups. Notably, the self-explanation and erroneous example groups improved in the mean number of optimal responses provided from the immediate to delayed post-test, while the control group worsened (Figure \ref{fig:overallbar}). Furthermore, the self-explanation group experienced the greatest decrease in regret, while the control group's cumulative regret increased (Figure \ref{fig:regbar}). This suggests that the interventions we have introduced, exposure to erroneous examples and prompting for open-ended self-explanation, improve retention of the policy over time. This finding echoes trends found in educational literature; across subject levels, students who work with self-explanations and erroneous examples exhibit greater concept retention than students who simply work through the problem (as in our control group) \cite{mclaren, errex, selfex1}. The similarities between our results and those in educational literature indicate the value in continuing to apply findings from traditional classroom learning to teaching robot policies.  

While we observe interesting trends in the change in performance between the immediate and delayed post-tests across the test conditions, mixed ANOVAS did not find significant effect of test condition on change in number of optimal responses ($F(1, 30) = 0.539, p = 0.590$) or on change in the sum of regret ($F(1, 30) = 1.533, p = 0.234$).

\section{Improved Understanding}
\textbf{H5} is not supported. A mixed ANOVA did not reveal a statistically
significant effect of test conditions on ratings of understanding
($F(1, 39) = 1.434, p = .252$). We hypothesized that the exposure to common misconceptions via erroneous examples would be informative to users, but this does not appear to be the case. In piloting, we observed that participants often rated tests as more informative when they confirmed the participant's current beliefs. Perhaps the erroneous examples often contradicted participants' beliefs and caused confusion, which participants then perceived as reducing their understanding rather than improving it by correcting their misconceptions. 

\section{Cognitive Load}
\textbf{H6} is not supported. A mixed ANOVA did not find significant effect of test condition on participants' self-reported cognitive load ($F(1, 39) = 1.076, p = 0.352$). This is an interesting result, because the self-explanation and erroneous explanation test conditions required additional work from participants (analyzing erroneous examples and providing open-ended explanations). We hypothesized that the increase in modalities of response required of the self-explanation and erroneous explanation groups would make participants feel that more effort was required. However, it is possible that the framing of the statement assessing cognitive load ``Learning the game strategy was taxing" did not prompt participants to consider the responses required of them, but just the policy instead. We also observed that some participants gave brief, and sometimes nonsensical, responses to explanation prompts; perhaps the effort participants spent was concentrated in providing the trajectories in new environments and little effort was spent on providing self-explanations or comparing their beliefs to erroneous examples. 

\section{Analysis of Explanation Responses}
The collection of open-ended self-explanation and explanation of error responses yielded a rich set of qualitative data that we use to identify the reasoning styles that learners adopted. We categorized explanations into five categories: imitation learning (IL), inverse reinforcement learning, procedural, understanding, and other. Responses falling into the imitation learning category were characterized by references to previous demonstrations and tests; explanations in this category indicated that participants were attempting to emulate what they had seen previously without thinking critically about rewards and tradeoffs. Responses in the inverse reinforcement learning category were characterized by references to features of the environments, their costs, and their tradeoffs. This indicated that the human learner was attempting to infer the underlying reward function. Responses in the procedural category made references to the task mechanics and instructions. For example, participants might explain themselves by saying they wanted to bring the agent to the destination or that they wanted to minimize energy loss without explaining \textit{how} they were doing so. Responses in the understanding category were statements about the participant's level of confusion or confidence about the task/policy. 

We had two coders categorize the explanation responses. There were 363 explanations to analyze in total, between the self-explanations and explanations of error. Each coder coded 220 responses, with an overlap of 72 responses. These responses were used to asses inter-coder reliability. We found that our coders had 78 \% agreement. We also assessed inter-coder reliability with Cohen's kappa to remove the effect of codings that agreed by chance. We found a Cohen's kappa of 0.63, which  can be interpreted as a moderate level of agreement \cite{reliability}. 
 
\begin{figure}[!htbp]
    \centering
    \includegraphics[height = 7 cm]{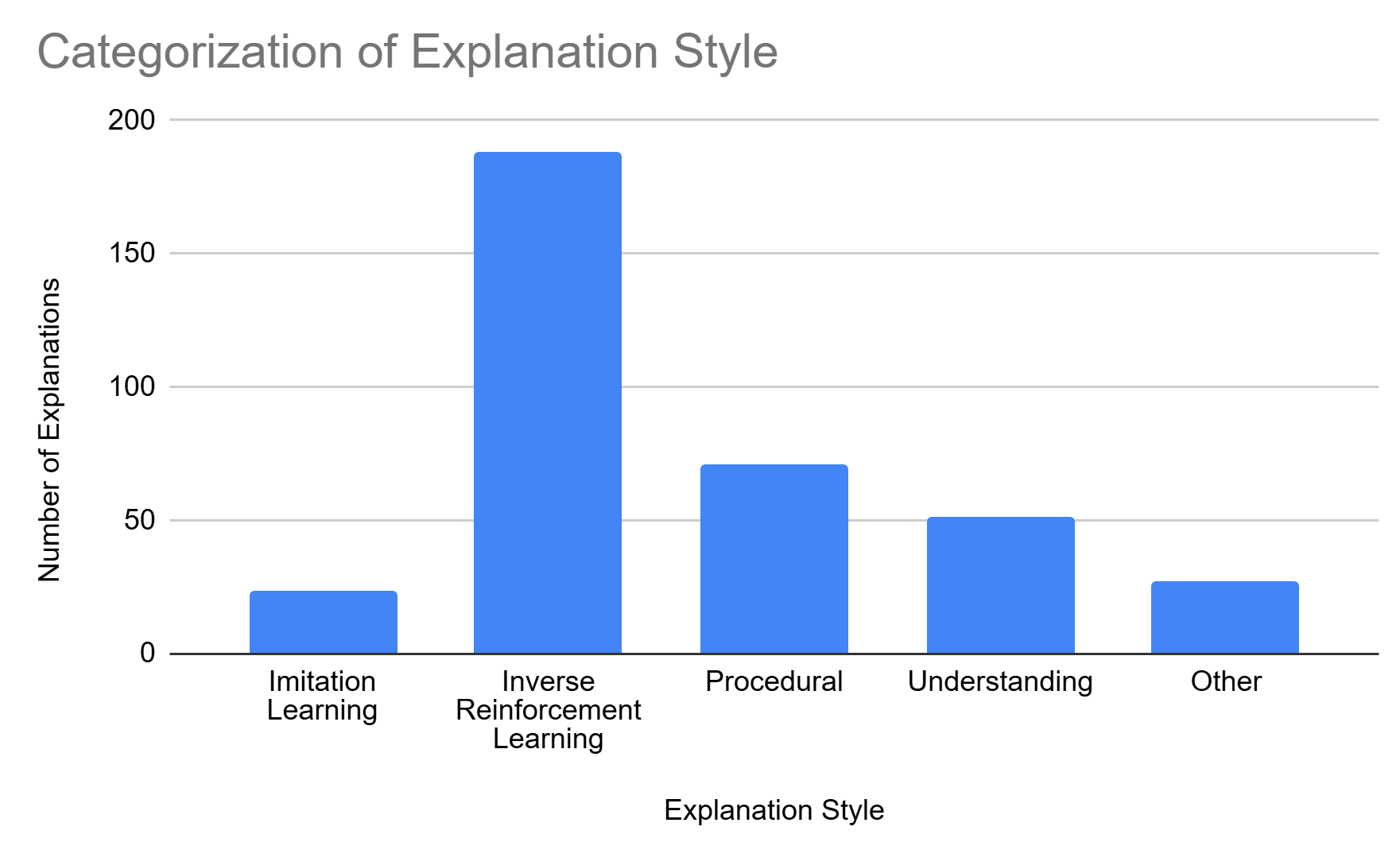}
    \caption{Prevalence of explanation types among self-explanation and erroneous example groups. }
    \label{fig:codings}
\end{figure}

\begin{figure}[!htbp]
    \centering
    \includegraphics[height = 8 cm]{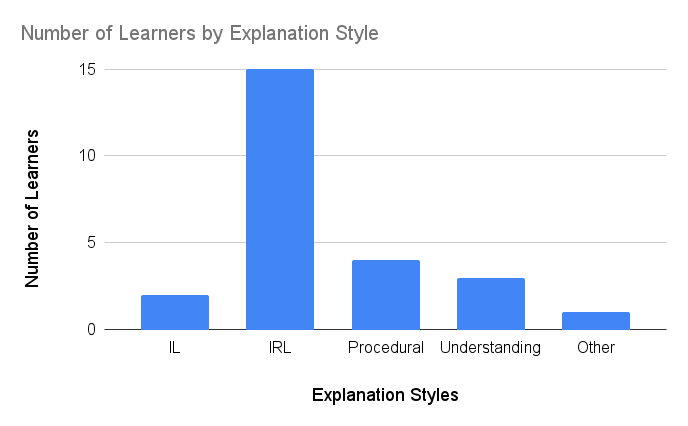}
    \caption{Categorization of participants in self-explanation and erroneous example groups by learning style. }
    \label{fig:style}
\end{figure}

As seen in Figure \ref{fig:codings}, IRL-based explanations were the dominant type of explanation. We additionally categorized participants into learning/explanation styles, based on the most frequent type of explanation they provided (Figure \ref{fig:style}). Most participants in the self-explanation and erroneous example groups were categorized as IRL learners; they seemed to focus most of their reasoning on rewards and costs associated with features in the environment, rather than copying what they had seen in demonstrations or simply re-stating the goal of the task. The prevalence of IRL-based reasoning validates assumptions made in this work and prior works that humans learning robot policies use an inverse reinforcement learning style to assess the agent's behavior. This is a novel contribution in itself; we are able to substantiate the assumptions made in modeling human beliefs over robot policies through our required explanation prompts. Previous studies have used optional survey questions to understand how learners may gain information from demonstrations and tests. However, data from these optional responses was far sparser than the explanations we collected and did not directly prompt the participants to explain their reasoning. 

\begin{figure}[!htbp]
    \centering
    \includegraphics[width=\linewidth]{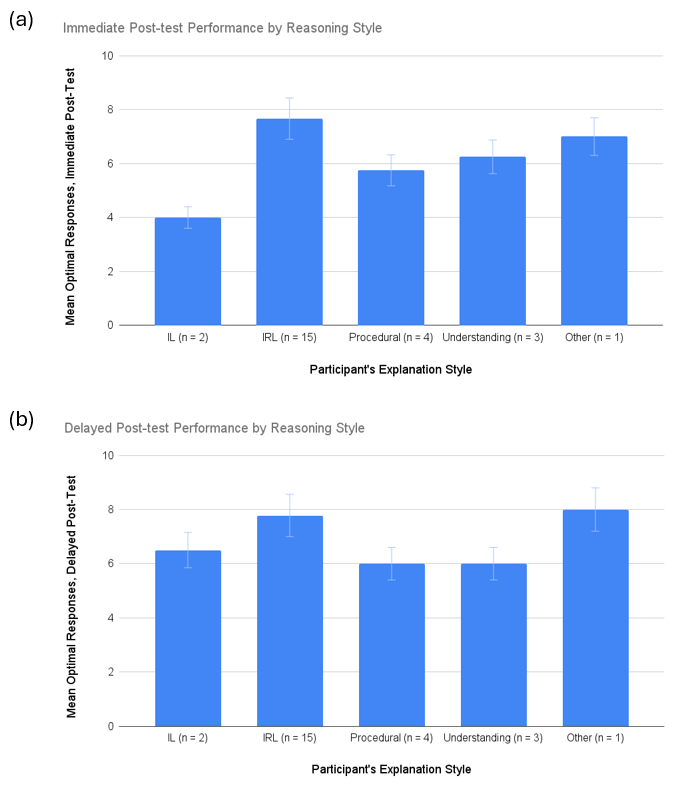}
    \caption{(a) Average number of optimal responses on immediate post-test by style. (b) Average number of optimal responses on delayed post-test by style. IRL-style learners perform better than most other groups.}
    \label{fig:style_opt}
\end{figure}

\begin{figure}[!htbp]
    \centering
    \includegraphics[width=\linewidth]{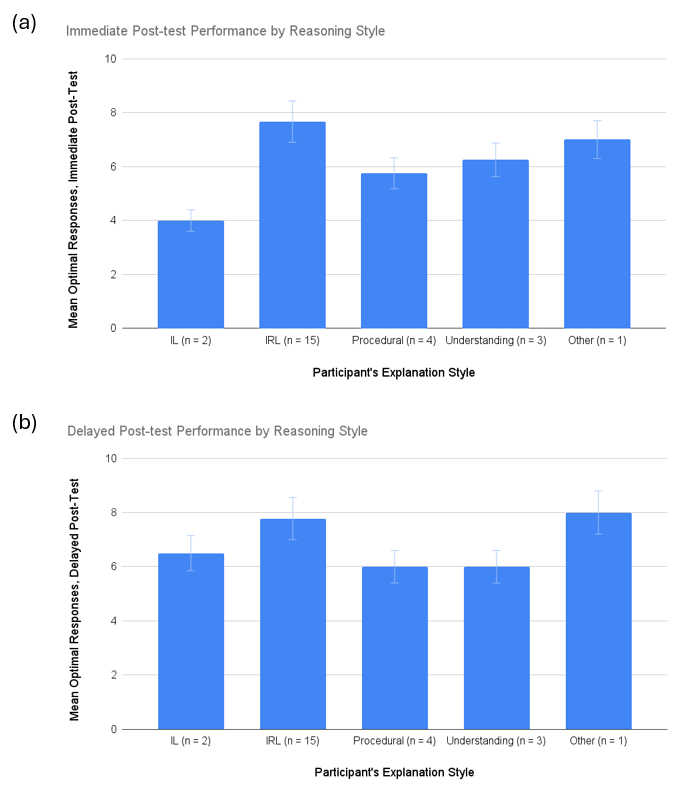}
    \caption{(a) Average regret on immediate post-test by style. (b) Average regret on delayed post-test by style. IRL-style learners perform better than most other groups.}
    \label{fig:style_reg}
\end{figure}

Following categorization of learners into different explanation styles, we compared performance on the immediate and delayed post-tests by style (Figure \ref{fig:style_opt}, Figure \ref{fig:style_reg}). We found that in both performance metrics, optimal response and regret, learners who most commonly reasoned through IRL outperformed their peers who used other reasoning styles. This comes with one exception: there was a single learner whose responses most commonly fell into the "other" category. This participant outperformed the other groups on some metrics, but was often similar in performance to the IRL learners. These findings, too, are promising. If researchers have assumed that learners are following IRL-like reasoning and tailoring their interventions to IRL models, it is critical that IRL reasoning be successful. We present evidence that this style of reasoning is natural to people learning reinforcement policies policies and that it enables participants to learn these policies better. It could also be the case that people who naturally lean towards IRL reasoning perform better at policy prediction tasks; in this case, we as roboticists should foster IRL reasoning in educating the public about robot behavior. 

\section{Discussion - Comparison with Prior Work}
Prior work has made a strong case for adapting techniques for teaching students in a classroom to teaching robot policies to human learners. Work by Lee et al. finds that scaffolding lessons improves learner performance in predicting robot actions, and in a follow ups study, Lee et al. find that incorporating feedback on unit tests into the learning process does the same  \cite{lee2022, lee3}. While we cannot definitively conclude that self-explanations and erroneous examples improve learner performance in predicting robot actions, we observe trends in the change in optimal response and regret between immediate and delayed post-tests that suggest these classroom techniques improve policy retention. We believe that the use of erroneous examples and self-explanation in combination with proven techniques such as scaffolding and incorporating feedback will improve both immediate information gain and policy retention. 

The works we have discussed iterated on the presentation of scaffolding and feedback in order to achieve their results and sampled from a  much larger participant pool; as such it is difficult to directly compare our work with these studies. However, we do see a significant interaction between domain and final test performance on both the immediate and delayed post-test, which was observed by Lee et al. as well \cite{lee3}. 

Through our analysis of self-explanations and explanations of error provided by learners in our study, we are also able to support the assumption in these prior works that humans learning reinforcement learning policies can be modeled using IRL. Participants frequently verbalized their reasoning in terms of the rewards or costs incurred by reaching different environment features. For example, in the taxi domain one learner said  ``Grabbing green would significantly increase the distance taken to get to the goal, so it is not worth it'' in reference to reaching the recharge station costing more energy than the station itself would give back. IRL-style participants also made references to an exploratory process of learning the rewards, explaining ``[I'm] testing my theory of green's value'' when attempting to determine the tradeoff between the recharge station and step cost. Participants exhibiting this style of reasoning appear to perform better than those making references to examples shown already, ``I simply followed the demonstration I was given previously, " or those restating the goal, ``Dropping off the circle at the grey square." We see promising use of IRL reasoning by participants, which can be used by researchers in the future to more accurately model human learning of robot policies and build educational modules based on these models. 

\chapter{Conclusion}
\section{Limitations}
Although the trends we observe in our results are promising, we were unable to find statistically significant differences between treatment groups across quantitative and qualitative metrics. This may be partially attributed to the small sample size in our user study, as well as the high variance in performance for the self-explanation and erroneous example treatment groups. This  high variance itself may be indicative of the treatment's effect on the learning process. In the future, the user study should be expanded in size; based on the effect size of similar studies teaching robot policies to human learners we would require at least 20 participants per group to see similar changes in policy prediction ability based on the intervention introduced \cite{lee3}. Based on studies of erroneous examples in traditional classroom learning, we may require as many as 40 participants per group \cite{mclaren}.
\section{Future Work}
The use of self-explanation and erroneous examples in robot policy demonstrations are  novel applications. As such, there are many directions yet to be explored with these pedagogical tools. For instance, erroneous examples may be integrated into the policy teaching loop as standalone demonstrations, rather than as part of unit tests. The format of the explanations (both self-explanations and explanations of error) could also be changed to multiple choice or fill-in-the-blank. This format shift may change the qualitative experience of the teaching loop (ex. may decrease cognitive load). Additionally, future researchers may vary the amount of time between the immediate and delayed post-tests or may introduce additional delayed post-tests; in the literature on erroneous examples in traditional classrooms, there is no standard time interval after which a concept is retained rather than immediately understood. When teaching different robot policies to humans, it may be more critical that a surgeon remembers the policy of an autonomous medical robot over the long term than a pedestrian remembers the policy of a package delivery robot.

Based on results regarding the interaction between treatment group and test difficulty, it may be interesting to examine the effect of self-explanation and erroneous examples on groups of learners with different prior knowledge. Educational literature has found that erroneous examples in math classrooms tend to benefit students with high prior knowledge more than those with low prior knowledge \cite{mclaren}. Perhaps roboticists should provide erroneous examples when deploying robots among groups familiar with autonomous agents,  but only show optimal demonstrations to novice learners. 

\section{Implications}
As robots become increasingly technologically capable, they will be integrated into homes and workplaces alongside humans at a greater pace. These people will come from various backgrounds and have a diverse set of opinions and knowledge about autonomous agents. Common to them, however, will be a need to understand the autonomous agents they will exist beside. Just as in human-human interaction, mutual understanding between humans and autonomous agents promotes trust between the groups and increases the safety of all involved \cite{endsley}. In this research, we work toward human understanding of robots by utilizing a robot policy teaching loop developed in literature and pedagogical practices of self-explanation and erroneous examples. In a user study incorporating the same, we observe trends of better policy retention as a result of these interventions. We also validate the research we build upon, observing similar effects of policy complexity on human learning and identifying IRL-like reasoning processes.  In improving robot policy retention among human learners, we build on prior work from other researchers and take further steps towards increasing transparency and trust in human-robot collaboration.

\bibliographystyle{ACM-Reference-Format}
\bibliography{bibtex}
\end{document}